\documentclass[]{elsarticle}
\usepackage[colorlinks=false,linkcolor=black, citecolor=black, urlcolor=black]{hyperref}

\usepackage{lineno,hyperref}
\modulolinenumbers[5]

\usepackage{tabularx} 
\usepackage{booktabs} 
\usepackage{amsmath}
\usepackage{multirow}
\usepackage{listings}
\usepackage{color}
\usepackage{subfig}
\usepackage{algorithm}
\usepackage{algpseudocode}
\definecolor{lightgray}{gray}{0.9}
\newcommand{\inlinecode}[1]{\colorbox{lightgray}{\lstinline|#1|}}

\journal{Arxiv}
\date{}

\begin{document}

\begin{frontmatter}

\title{CodeGRU: Context-aware Deep Learning with Gated Recurrent Unit for Source Code Modeling}






\author[add1]{Yasir Hussain\corref{cor1}}
\ead{yaxirhuxxain@nuaa.edu.cn}
\author[add1,add2,add3]{Zhiqiu Huang\corref{cor1}}
\ead{zqhuang@nuaa.edu.cn}
\author[add1]{Yu Zhou}
\ead{zhouyu@nuaa.edu.cn}
\author[add1]{Senzhang Wang}
\ead{szwang@nuaa.edu.cn}

\address[add1]{College of Computer Science and Technology, Nanjing University of Aeronautics and Astronautics (NUAA), Nanjing 211106, China}
\address[add2]{Key Laboratory of Safety-Critical Software, NUAA, Ministry of Industry and Information Technology, Nanjing 211106, China}
\address[add3]{Collaborative Innovation Center of Novel Software Technology and Industrialization, Nanjing 210093, China}

\cortext[cor1]{Corresponding author}

\begin{abstract}
	
	Context: Recently deep learning based Natural Language Processing (NLP) models have shown great potential in the modeling of source code. However, a major limitation of these approaches is that they take source code as simple tokens of text and ignore its contextual, syntactical and structural dependencies.
	
	Objective: In this work, we present CodeGRU, a gated recurrent unit based source code language model that is capable of capturing source code’s contextual, syntactical and structural dependencies.
	
	Method: We introduce a novel approach which can capture the source code context by leveraging the source code token types. Further, we adopt a novel approach which can learn variable size context by taking into account source code’s syntax, and structural information.
	
	Results: We evaluate CodeGRU with real-world data set and it shows that CodeGRU outperforms the state-of-the-art language models and help reduce the vocabulary size up to 24.93\%. Unlike previous works, we tested CodeGRU with an independent test set which suggests that our methodology does not requisite the source code comes from the same domain as training data while providing suggestions. We further evaluate CodeGRU with two software engineering applications: source code suggestion, and source code completion.
	
	Conclusion: Our experiment confirms that the source code’s contextual information can be vital and can help improve the software language models. The extensive evaluation of CodeGRU shows that it outperforms the state-of-the-art models. The results further suggest that the proposed approach can help reduce the vocabulary size and is of practical use for software developers.

\end{abstract}


\end{frontmatter}


\section{Introduction \label{Introduction}} 
Source code suggestions, code completion, bug fixing, etc. are vital features of a modern integrated development environment (IDE). These features help software developers to build and debug software rapidly. In the last few years, there have been a massive amount of increase in code related databases over the internet. Many open source websites (i.e. w3school, GitHub, Stack Overflow, etc.) provides API libraries, code usage examples, bug fixing, and much more. Software developers exceedingly rely on such resources for above-mentioned purposes.

Natural language processing (NLP) \cite{cambria2014jumping,manning2014stanford,ranjan2016survey} explores, understands and manipulates natural language text or speech to do serviceable things. NLP techniques have shown its effectiveness in many fields such as speech recognition \cite{bellegarda2000large}, information retrieval \cite{berger2017information}, text mining \cite{rajman1998text}, machine translation \cite{xiao2018machine} and source code modeling \cite{hindle2012naturalness,raychev2014code,karaivanov2014phrase,tu2014localness}. One of the most common NLP technique for source code modeling is statistical language models (SLM), which calculates the probability distribution over sequences in a corpus. Given a sequence $S$ of length $N$ it assigns the probability to the whole sequence $Pb(t_1,....t_n)$ and then calculates the likelihood of all sub-sequences to find the most likely next sequence.

The advancement in the neural network based NLP models \cite{white2015toward,raychev2014code,young2018recent} have recently shown that they can effectively overcome the context issue that cannot be effectively addressed by SLM \cite{hindle2012naturalness,tu2014localness} based models. Many deep learning based approaches have been applied for different tasks for source code modeling such as code summarization \cite{iyer2016summarizing,allamanis2016convolutional},  code readability classification \cite{mi2018improving}, code generation \cite{sethi2018dlpaper2code}, error fixing \cite{gupta2017deepfix,xiao2019improving}, and code recommendation \cite{white2015toward,raychev2014code,gu2016deep,dam2016deep}. A major limitation of these approaches \cite{hindle2012naturalness,white2015toward,raychev2014code} is that they take source code as simple tokens of text sequence and ignore its contextual, syntactical and structural dependencies. Another limitation is that they learn source code as a sequence to sequence problem with fixed size context where the right context may not be captured in the fixed size window, which leads to the inaccurate prediction of the next code token.

Compared with natural language text, source code tends to have richer contextual, syntactical and structural dependencies. Treating source code as a simple text cannot effectively capture these dependencies. Software developers usually choose to have different names for methods, classes, and variables, which makes it difficult to capture the right context. For example, one software developer may choose an identifier name \inlinecode{num} for an \inlinecode{INT} data type, while another one may choose \inlinecode{size} for the same purpose. Consider another example where a common method \inlinecode{i.tostring()} converts a variable to \inlinecode{String} data type. A similar method \inlinecode{person.tostring()} refers to an object of a $person$ class that returns a person’s information. In addition, the source code must follow the rules defined by its grammar. For example, a \inlinecode{try} block must be followed by a \inlinecode{catch} block. Another example is that when a developer uses \inlinecode{do} block, the next block should be \inlinecode{while()} and the possible next token suggestion should be \inlinecode{;} according to the syntax of java language grammar.  Previous works \cite{hindle2012naturalness,white2015toward,raychev2014code} lack to consider such information which can be valuable for source code modeling tasks.

In this work, we present CodeGRU which considers the source code's context, syntax, and structure while suggesting the next source code token. This work does not simply consider source code as simple tokens of text. The CodeGRU introduces a novel approach which can correctly capture the source code context by leveraging the token type information. The CodeGRU can effectively capture the right context even it is separated far apart in the code. CodeGRU further proposes a novel approach which can learn variable size context while modeling source code. Unlike previous works \cite{hindle2012naturalness,white2015toward,raychev2014code,nguyen2018deep}, we do not treat the source code as a single sequence of text tokens instead we use a novel approach which builds the sequences based on source code’s structural and syntactical information. We evaluated CodeGRU with real-world data set with an independent test set with two software engineering applications: source code suggestion, and source code completion.

This work makes the following unique contributions:
\begin{itemize}
	\item A novel approach for source code modeling is proposed, which can capture the source code context by leveraging the token type information. 
	\item A novel method which learns the variable size context of the source code is proposed. Unlike previous works, we do not treat the source code as a single sequence of text tokens instead we use a novel approach which builds the sequences based on source code’s structural and syntactical information. 
	\item An extensive evaluation of CodeGRU on the real-world data set shows that CodeGRU outperforms the state-of-the-art language models. We further evaluated CodeGRU with two software engineering applications: (1) source code suggestion, which can suggest multiple predictions for the next code token, and (2) code completion, which can complete the whole next code sequence. 
	
\end{itemize}

\section{Preliminaries \label{Technical}} 
In this section, we will discuss the preliminaries and technical overview of this work.

\begin{figure}[htb]
	\centering
	\includegraphics[width=0.35\linewidth]{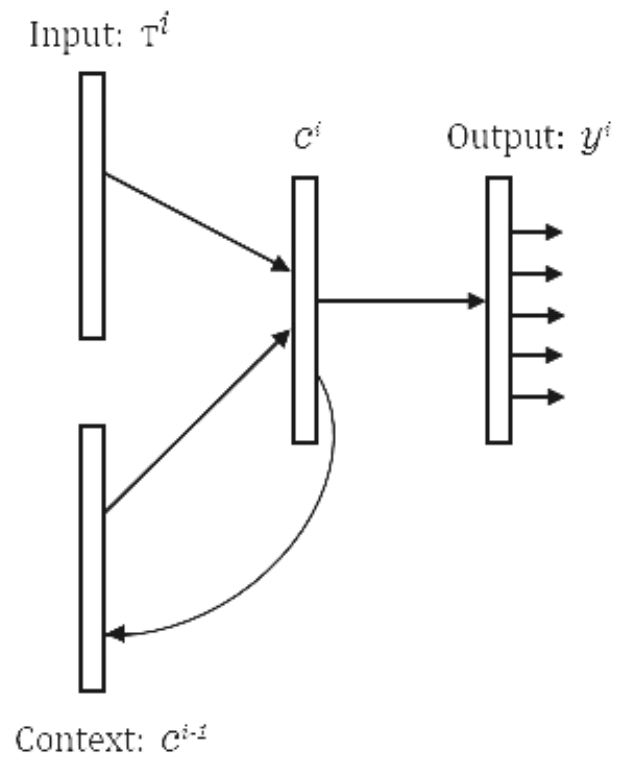}
	\caption{An architecture of a RNN neuron where input is a code token vector at index $i$, and the outputs are different next code tokens $y^i$ based on the context and probabilities}
	\label{fig:RNNModelArchitecture}
\end{figure}

\hyperref[fig:RNNModelArchitecture]{Fig.} \ref{fig:RNNModelArchitecture} shows the architecture of the RNN for source code modeling, where $\tau$ is input layer, $c$ is context layer also known as hidden layer and $y$ is the output layer. The hidden state activation at a time step $i$ is computed as a function on the previous $h_{i-1}$ along with current code token $\tau_i$.
\begin{equation}
h_i = f(\tau_i,h_{i-1})
\end{equation}

Usually $f$ is composed of an element-wise nonlinear and affine transformation of $\tau_i$ and $h_{i-1}$.

\begin{equation} \label{eq:2}
h_i = \phi(W\tau_i,Uh_{i-1})
\end{equation}

Here $W$ is the weight matrix for the input to hidden layer and $U$ is the weight matrix for the state to state matrix, and $\phi$ is an activation function. The RNN models \cite{mikolov2010recurrent,funahashi1993approximation} tends to look back further than $n-1$. But vanilla RNN suffers from vanishing gradient problem which can be overcome by using Gated Recurrent unit (GRU) model.

The GRU exposes \cite{young2018recent} the full hidden content without any control which is ideal for source code modeling. It is composed of two gates, the rest gate $r_i$ and the update gate $z_i$. Further, it entirely exposes its memory context at each time step $i$. Exposing the entire context on each time step helps to learn contextual dependencies better than vanilla RNN. It can be expressed as
\begin{equation}
h_i = (1-z_i)h_{i-1}+z_i\hbar_i
\end{equation}

Where $h$ and $\hbar$ is prior context and fresh context respectively.

\begin{equation}
z_i = \phi(W_z\tau_i+U_zh_{i-1})
\end{equation}

\begin{equation}
\hbar_i = tanh(W\tau_i+r_i \otimes Uh_{i-1})
\end{equation}

\begin{equation}
r_i = \phi(W_r\tau_i+U_rh_{i-1})
\end{equation}
A major difference from \hyperref[eq:2]{Eq.} \ref{eq:2} is that the $\hbar$ is modulated by the reset gates $r_i$. Here $\otimes$ is element-wise multiplication and $\phi$ is the activation function.

\section{The CodeGRU Model \label{CodeGRUInDetail}}
\begin{figure*}[h]
	\centering
	\includegraphics[width=\linewidth]{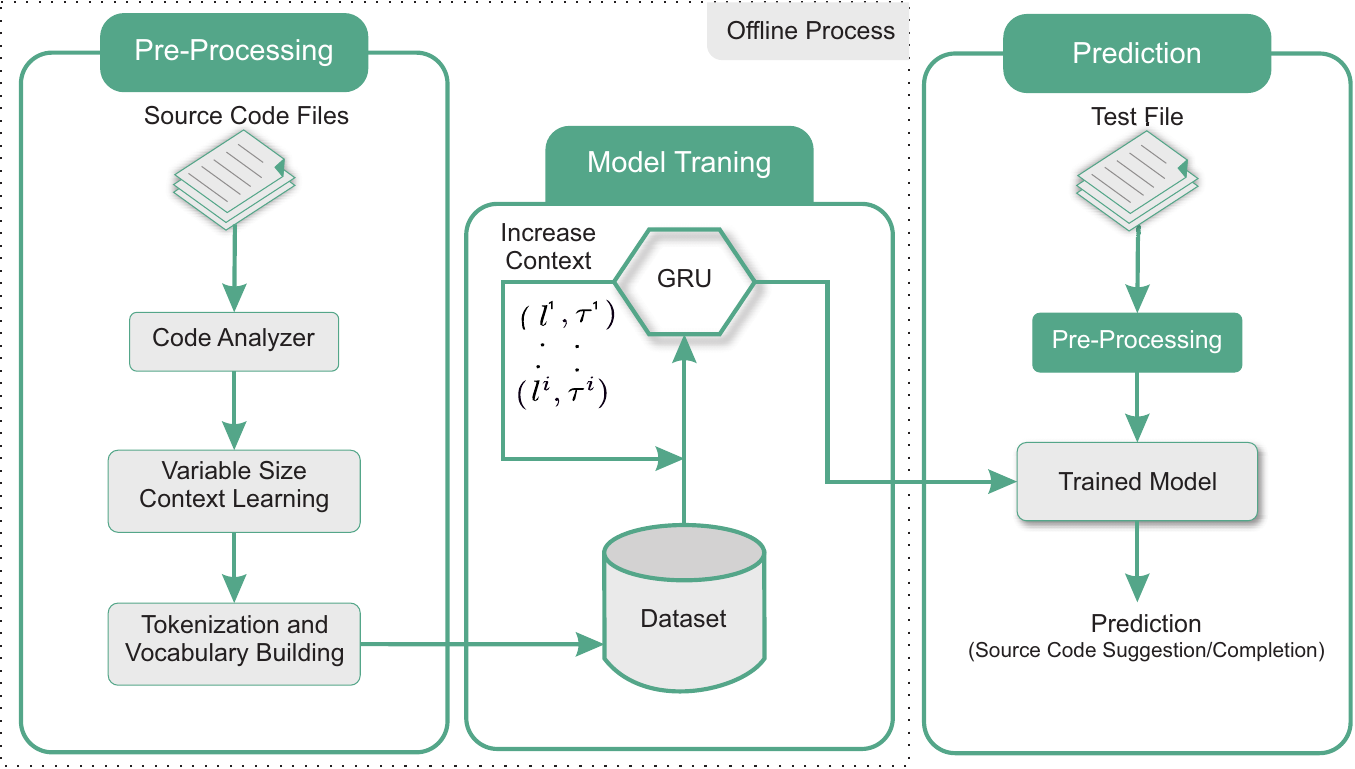}
	\caption{The framework of CodeGRU, which is a context aware deep learning model for source code modeling.}
	\label{fig:CodeGRUModelArchitecture}
\end{figure*}

In this section, we introduce CodeGRU in detail. The overall workflow of CodeGRU is illustrated in \hyperref[fig:CodeGRUModelArchitecture]{Fig.} \ref{fig:CodeGRUModelArchitecture}. The first step is data collection, which we will discuss in \hyperref[Dataset]{section} \ref{Dataset}. Next step is \textit{Code Analyzer}, which pre-processes the source code files and captures the token type information. The \textit{Code Analyzer} parses the source code and encode the type information to capture the source code context. Next, the \textit{Variable Size Context Learning} approach is used to build the sequences based on the source code syntax and structure. Next, we tokenize the sequences and build vocabulary. Finally, we train deep learning classifier for source code suggestion and completion task. Each step is discussed in detail in the following subsections.

\subsection{Code Analyzer \label{CodeAnalyzer}} 
Code Analyzer first takes a source code file and pre-process it to capture the source code context. First, we normalize the files by removing all blank lines, inline and block level comments as they have no impact on source code suggestion or completion task. Source code consists of different kinds of tokens such as classes, functions, variables, literals, language-specific keywords, data types, stop words, etc. Among all these, language dependent keywords, stop words, library functions, and data types form a shared vocabulary which can be considered as context. A key insight is that to capture the context of source code tokens we capture their token types. Here we care about the token type rather than the token identifier. As discussed earlier code token identifiers can differ from developer to developer, so we use the token types to help capture the context information. The exact values of literals (\textit{Int, float, long, double, byte, String, etc}) are unnecessary in source code modeling task. So, we transform all literal values to their abstract data types according to the Java language grammar\footnote{https://docs.oracle.com/javase/specs/jls/se7/html/jls-18.html}. For example given \inlinecode{System.out.println("Hello World")}, the string value \textit{"Hello World"} is of type \inlinecode {String Literal} according to java language grammar. We transform it with its token type \inlinecode { StringLiteral}. Similarly, the value of identifier \textit{a} in \inlinecode{a = 1.1} is not important, so we transform \textit{1.1} with its token type \inlinecode {FloatLiteral}.

\begin{figure*}[h]
	\centering
	\includegraphics[width=0.5\linewidth]{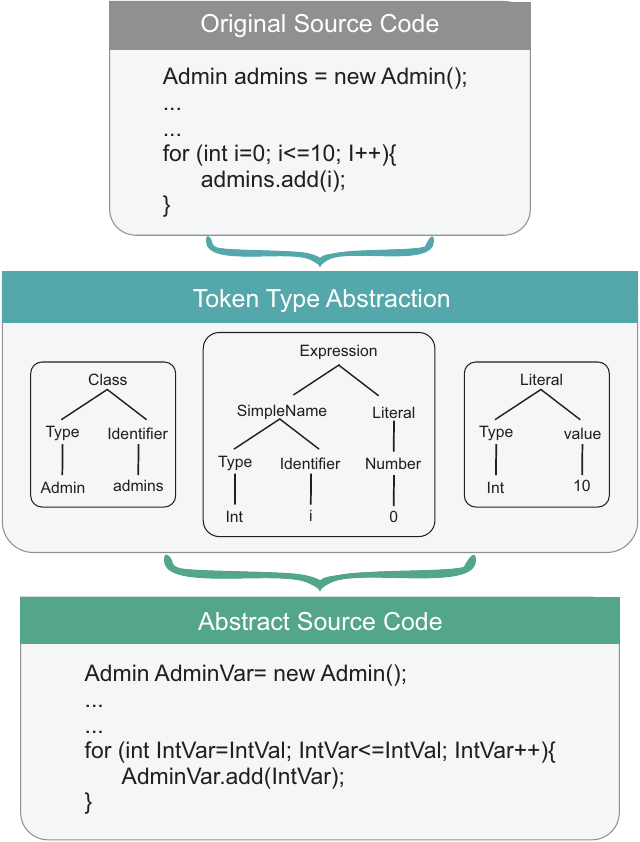}
	\caption{An example of Java code with our \textit{Code Analyzer} approach.}
	\label{fig:CodeAnalyzer}
\end{figure*}

A challenging issue in source code modeling is to capture the variable and class object identifies declaration types. Static type languages (Java, C++, C\#, etc.) are strongly type defined languages, which means types of such declarations need to be defined before use. In \hyperref[fig:CodeAnalyzer]{Fig.} \ref{fig:CodeAnalyzer} one can see a transformation example of variable and class object identifiers. We capture all identifiers types into their declared data types. In \hyperref[fig:CodeAnalyzer]{Fig.} \ref{fig:CodeAnalyzer} one can see, we transform all instances of identifier \textit{i} with its declared data type \inlinecode {Int} combined with a special token \inlinecode{Var}. Similarly, the class object identifier \textit{admins} is replaced with \inlinecode {AdminVar}. We leave special code tokens (\textit{true, false, null}) unchanged. Unlike literals, variables, and class objects, such tokens reflect constant behavior, which does not need transformation. \hyperref[Table:CodeAnalyzer]{Table} \ref{Table:CodeAnalyzer} shows some common code tokens and their resolved types captured by our novel approach.

\begin{table*}[h]
	\tiny
	\caption{Common source code examples by our \textit{Code Analyzer} approach}
	\label{Table:CodeAnalyzer}
	\begin{center}
		\setlength{\tabcolsep}{12pt}
		\resizebox{\textwidth}{!}{%
			\begin{tabular}{*{3}{r}}
				\toprule
				{Code Token} & {Token Type} & {Special Token}\\
				\midrule
				$i$ & Int & IntVar\\
				$"Hello World"$ & String Literal & StringLiteral\\
				$null$ & Literal & NullLiteral\\
				$outFile$ & File& FileVar\\
				$ex$ & Exception & ExceptionVar\\
				$arr$ & ArrayList<String> & ArrayListStringVar\\
				$lstID$ & List<Int>& ListIntVar\\
				$'c'$ & char & CharLiteral\\
				$true$ & Boolean & true\\
				$Int * x$ & Int Pointer & Int * IntPointerVar \\
				$Int ** x$ & Int Pointer & Int ** IntPointerToPointerVar \\
				$true$ & Boolean & true\\
				$inputfile.open()$ & File & FileVar.open()\\
				\bottomrule
		\end{tabular}}
	\end{center}
\end{table*}

\subsection{Variable Size Context Learning \label{ICSGen}} 
\begin{figure*}[h]
	\centering
	\includegraphics[width=0.5\linewidth]{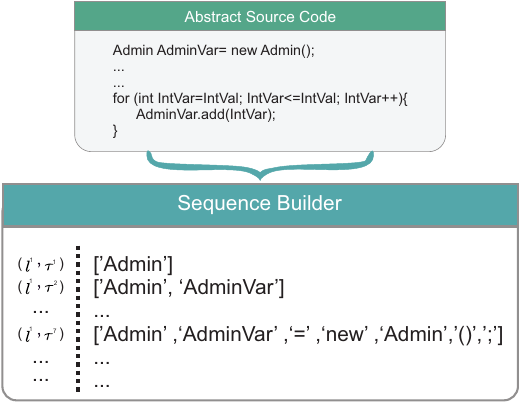}
	\caption{An example of variable size context learning approach.}
	\label{fig:VSCL}
\end{figure*}

Programming languages strictly follow the rules defined by their grammar. Each line in a programming language starts with a language reserved identifier, variable or class object deceleration, assignment statements, etc. Whereas an assignment statement can only have a variable name, object instance or array index on the left-hand side. Further, source code follow block rules such as \inlinecode{try-catch-final}, \inlinecode{do-while} where one must follow the other. We use such information while building the sequences. Unlike previous works \cite{white2015toward,raychev2014code,nguyen2018deep}, we do not consider the source code as a single sequence of text tokens and divide them into fixed-size context window. Instead, we leverage from the syntactical and structural information to learn variable size context of the source code. The CodeGRU takes a source code program and split it based on a code statement or a block statement. In source code a single statement ends with \inlinecode{\;} token where a block statement starts with the \inlinecode{\{} token and ends with \inlinecode{\}} token according to the Java language grammar\footnote{https://docs.oracle.com/javase/specs/jls/se7/html/jls-18.html}. With this approach, we split each file into multiple sequences of code tokens \textit{X}. Here the goal is to produce the next token \textit{y} by satisfying the context of \textit{X}. We can express a source code file \textit{S} at line \textit{L}. Then a source code program can be represented as \textit{$(l^i,\tau^i)$} where $l^i$ is the line number and $\tau^i$ is tokenization of \textit{S} at $l^i$. It breaks each \textit{$l^i$} into several $\tau^i$ by iteratively increasing the context on each iteration. The CodeGRU learns the source code context at \textit{$(l^i,\tau^i)$} and keeps increase the \textit{$\tau^{i+1}$} until it reaches the upper bound limit of \textit{$l^i$}. When CodeGRU reaches the upper bound limit of \textit{$l^i$}, it increases \textit{$l^{i+1}$} and keeps learning the source code context. \hyperref[fig:VSCL]{Fig.} \ref{fig:VSCL} shows an example of proposed variable size context learning approach.

Further, here we expect the model to assign the high probability to the correct next source code suggestion by having a low \textit{Cross-entropy}. The \textit{Cross-entropy} is a cost function to observe how best the model works. A low value of \textit{Cross-entropy} indicates a good model. It can be expressed as
\begin{equation}
H(C) \approx -\dfrac{1}{N}\sum_{i=1}^{N}log_2 \text{\textit{ Pb}}_{C}(\tau^i|\tau^{i-1}_{i-n+1})
\end{equation}

\subsection{Tokenization and Vocabulary Building} \label{VocabBuild}
\begin{figure*}[htb]
	\centering
	\includegraphics[width=0.5\linewidth]{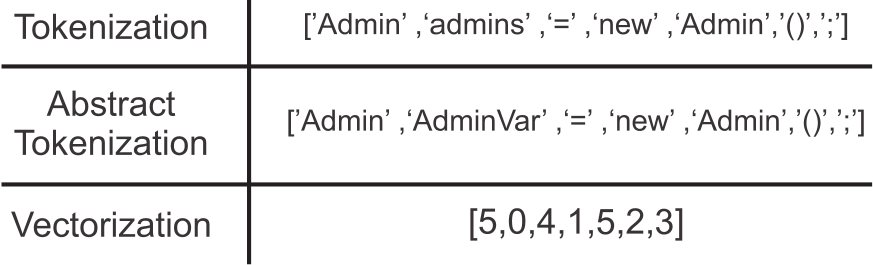}
	\caption{Process of building vocabulary for source code language models.}
	\label{fig:VocabBuild}
\end{figure*}

To convert the source code files into a form that is suitable for training we perform a series of transformations. First, we tokenize the source code files as shown in \hyperref[fig:VocabBuild]{Fig.} \ref{fig:VocabBuild}. Each unique source code token corresponds to an entry in the vocabulary. Each source code token is then assigned a unique positive integer value. In \hyperref[Table:VocabBuild]{Table} \ref{Table:VocabBuild} one can see the vocabulary statistics, where $V_{Norm}$ shows the vocabulary statistics without \textit{Code Analyzer} and {$V_{CodeGRU}$} shows the vocabulary statistics with \textit{Code Analyzer} approach. The vocabulary statistics without \textit{Code Analyzer} is reported by calculating the unique code tokens found in source code files after removing blank lines, inline and block level comments. Further, we replace all literal values to their abstract data types and leaving the rest of the file to its original state. 

\begin{table*}[h]
	\tiny
	\caption{Vocabulary statistics between projects.}
	\label{Table:VocabBuild}
	\begin{center}
		\setlength{\tabcolsep}{12pt}
		\resizebox{\textwidth}{!}{%
			\begin{tabular}{*{8}{r}}
				\toprule
				{} & {Min} &{Max} &{Mean} & {Median} & {S.D.} & {Total}\\
				\midrule
				{$V_{Norm}$} & 6813 & 47756 & 20742.7 &  19699.5 & 11579.53 & 177,342\\
				{$V_{CodeGRU}$} & 5566 & 33867 & 15358.7 &  13733.5 & 8503.74 & 133,135 \\
				\bottomrule
			\end{tabular}
		}
	\end{center}
\end{table*}

\subsection{Training and Prediction} \label{Pseudo}
We parse each source code file in the training set and use \textit{Code Analyzer} to capture the token type information as discussed earlier in \hyperref[CodeAnalyzer]{Section} \ref{CodeAnalyzer}. After normalizing and type encoding the source code files we use the variable size context approach to generate the sequences based on source code's syntax and structure as discussed in \hyperref [ICSGen]{Section} \ref{ICSGen}. To convert the sequences into a form that is suitable for training and to build vocabulary, the sequences are converted into vector form as discussed in \hyperref [VocabBuild]{Section} \ref{VocabBuild}. We use these sequences as input to CodeGRU for model training. The trained model is then used for the test purpose. For the prediction of next source code token \textit{y} in a file \textit{S}, we capture the token type information using \textit{Code Analyzer} prior to the prediction position \textit{y}. Then, we use the captured type information as context and input it to the model for the next source code suggestion. Then, each code token in the \textit{Vocabulary} is computed and ranked for the possible next source code suggestion.

\section{Empirical Evaluation \label{Experiments}}
In this section, we provide an empirical evaluation of CodeGRU. We train and evaluate our models on Intel(R) Xeon(R) CPU E5-2620 v4 with 16 cores running at \textit{2.10GHz} with \textit{64GB} of ram equipped with four NVIDIA Tesla K20m GPUs running CentOS v7 operating system. It took 8 days to fully train and test all models. One important thing to mention here is that the training and testing are offline and thus have no impact on prediction time. It takes less than 30 milliseconds for source code suggestion and source code completion tasks.

To evaluate the performance of the proposed approach, we aim at answering the following research questions:

\begin{itemize}
	\item RQ1: Does the proposed approach outperform the state-of-the-art approaches? if yes, to what extent?
	\item RQ2: How well does the proposed approach perform in source code suggestion and source code completion tasks?
	\item RQ3: To what extent the proposed approach helps reduce the vocabulary size?
\end{itemize}

To answer the research question (RQ1), we compare the performance of the proposed approach with the state-of-the-art approaches \cite{white2015toward,hindle2012naturalness,nguyen2018deep}. To answer the research question (RQ2), we evaluate and compare CodeGRU for source code suggestion and completion tasks with state-of-art approaches \cite{white2015toward,hindle2012naturalness,nguyen2018deep}. To answer the research question (RQ3), We provide the statistical results of vocabulary with and without our proposed approach. We further evaluate CodeGRU with two software engineering applications: source code suggestion, and source code completion tasks which show CodeGRU is of practical use.

\subsection{Dataset\label{Dataset}} 
To build our code database\footnote{\url{https://github.com/yaxirhuxxain/Source-Code-Suggestion}}, we collect open-source java projects from GitHub. We download an archive containing the latest snapshot of the project’s default branch. For comparison, we choose the projects used in previous studies \cite{hindle2012naturalness,nguyen2018deep,white2015toward} summarized in \hyperref[Table:DataSet]{Table} \ref{Table:DataSet}. The \hyperref[Table:DataSet]{Table} \ref{Table:DataSet} shows the version of the projects, total number of code lines, total code tokens and unique code tokens found in each project. We randomly choose one project(Cassandra) as an independent test set and use the rest of the projects for the model training purpose. Further, to evaluate the effectiveness of the proposed approach we adopt a random testing approach. In random testing we train the model on one project and test it with a different project (beside Cassandra and itself). To empirically evaluate our work, we repeat our experiment on each project separately. Each project is subdivided into ten equal lines of code folds from which one fold is used for validation and rest are used for training purpose. One important thing to mention here is that the test sets was never used while training or validating the models and was only used for evaluation purpose.

\begin{table*}[h]
	\tiny
	\caption{List of java projects used for evaluation. The table shows name of the project, version of the project, line of code (LOC), total code tokens and unique code tokens found in each project.}
	\label{Table:DataSet}
	\begin{center}
		\setlength{\tabcolsep}{12pt}
		\resizebox{\textwidth}{!}{%
			\begin{tabular}{*{6}{r}}
				\toprule
				&  &  &  \multicolumn{2}{c}{Code Tokens}\\
				{Projects} & {Version} &{LOC} &{Total} & {Unique}\\
				\midrule
				ant & 1.10.5 & 149,960 & 920,978 &  17,132\\
				cassandra & 3.11.3 & 318,704 & 2734218 &  33,424\\
				db40 & 7.2  & 241,766 & 1,435,382 &  20,286\\
				jgit & 5.1.3  & 199,505 & 1,538,905 &  20,970\\
				poi & 4.0.0  & 387,203 & 2,876,253 &  47,756\\
				maven & 3.6.0  & 69,840 & 494,379 &  8,066\\
				batik & 1.10.0  & 195,652 & 1,246,157 &  21,964\\
				jts & 1.16.0  & 91,387 & 611,392 &  11,903\\
				itext & 5.5.13 & 161,185 & 1,164,362 &  19,113\\
				antlr & 4.7.1 & 56,085 & 407,248 &  6,813\\
				\bottomrule
		\end{tabular}}
	\end{center}
\end{table*}

\subsection{Training \label{Training}} 
We train several baseline models for the evaluation of this work. In this section, we briefly describe the baselines for comparison in detail. We train the N-gram model used in Hindle et al. \cite{hindle2012naturalness}. We train the RNN \cite{raychev2014code} model used in White et al. \cite{white2015toward}. We train the DNN model used in Nguyen et al. \cite{nguyen2018deep}. We implement our own version of DNN since it is not publicly available. In our reproduction, we follow the process described in Nguyen et al. \cite{nguyen2018deep}. Further, we train four different models N-gram+, RNN+, GRU, and CodeGRU by using our proposed approach. The N-gram+ and RNN+ are trained similar to the previous studies by adopting our proposed \textit{Code Analyzer} approach. The GRU based model is trained by using our proposed approach, but without using the variable size context learning approach. The CodeGRU model is trained by using our proposed approach with variable size context as described in \hyperref[CodeGRUInDetail]{Section} \ref{CodeGRUInDetail}. For the training purpose, each source code file is tokenized as discussed earlier in \hyperref[VocabBuild]{Section} \ref{VocabBuild}. Then, we map the vocabulary to a continuous feature vector of dense size \textit{300} similar to Word2Vec \cite{rong2014word2vec}. This approach helps us build a dense vector representation for each vocabulary index without compromising over the semantic meaning of the source code tokens.

The \hyperref[Table:DeepModelsArch]{Table.} \ref{Table:DeepModelsArch} shows the architecture of deep learning based models. We choose \textit{300} hidden units with context size \textit{20}. We use \textit{Adam} \cite{kingma2014adam} optimizer with the learn rate set to its default \textit{0.001}. To control over fitting we use \textit{Dropout} \cite{gajbhiye2018exploration} at the rate of \textit{0.25}. We employ \textit{early stopping} \cite{alon2019code2vec,santos2018syntax} to help stop model training when it achieves the best performance.

\begin{table*}[h]
	\caption{Deep learning models architecture summary.}
	\label{Table:DeepModelsArch}
	\begin{center}
		\setlength{\tabcolsep}{12pt}
		\resizebox{\textwidth}{!}{%
			\begin{tabular}{*{6}{r}}
				\toprule
				& Type & Size & Activations  \\
				\midrule
				Input & Code embedding & 300 &  &  \\
				Estimator & RNN,GRU & 300 & tanh & \\
				Over Fitting & Dropout & 0.25 &  & \\
				Output & Dense & $V$ & softmax  & \\
				Loss & Categorical cross entropy& &  & \\
				Optimizer & Adam & 0.001 &  & \\
				\bottomrule
		\end{tabular}}
	\end{center}
\end{table*}

\subsection{Prediction \label{Pridiction}} 
For the prediction of next source code token \textit{y} in a source code file \textit{S}, the model takes the context information \textit{$(l^i,\tau^i)$} prior to the prediction position \textit{y}. We use the \textit{CodeAnalyzer} to capture the token's context information as discussed in section \ref{CodeGRUInDetail}. We then use the trained model to predict the most likely top-k suggestions for the given context. CodeGRU predicts the token type for an identifier and the actual token for rest of the code tokens. We use the top-k accuracy and Mean Reciprocal Rank (MRR) metrics for the evaluation of this work. 

\section{Results} \label{Results}

\subsection{Accuracy Comparison\label{Accuracy}} 
For comparison, we evaluate the models with \textit{top-k} accuracy score as done in the previous works \cite{hindle2012naturalness,white2015toward,raychev2014code,nguyen2018deep}. We calculate top-k accuracy, where \textit{k=1,3,5,10}. The accuracy scores for independent test (Cassandra) of all models are shown in \hyperref[Table:Accuracy]{Table} \ref{Table:Accuracy}. One can see that CodeGRU model outperforms other baseline \cite{hindle2012naturalness,white2015toward,raychev2014code,nguyen2018deep} models. We can see in \hyperref[Table:Accuracy]{Table} \ref{Table:Accuracy} that our proposed CodeGRU achieves the max accuracy score of \textit{46.74 @ k=1} and \textit{74.32 @ k=10}, whereas previous models gains much lower score \textit{39.26 @ k=1} and \textit{69.12 @ k=10}. Further, we can observe that our proposed approach help improve the performance of previous studies. The simple N-gram model achieves the max score of \textit{21.62 @ k=1} and \textit{31.62 @ k=10} where as with our proposed approach the N-gram+ achieves the max score of \textit{25.57 @ k=1} and \textit{37.84 @ k=10}. Similarly, by using our proposed approach the RNN+ achieves the max score of \textit{38.90 @ k=1} and \textit{70.09 @ k=10} where as the RNN model gains the max score of \textit{37.78 @ k=1} and \textit{67.63 @ k=10}.

\begin{table}[htbp]
	\caption{Accuracy comparison of proposed approach with previous works with independent test set (Cassandra).}
	\begin{center}
		\begin{tabular}{|c|c|ccc|cccc|}
			\toprule
			&       & \multicolumn{3}{c|}{Previous Works} & \multicolumn{4}{c|}{\textbf{Our Work}} \\
			& K     & N-gram & RNN   & DNN   & {N-gram+} & {RNN+} & {GRU} & {CodeGRU} \\
			\bottomrule
			\multirow{4}[2]{*}{antlr} & 1     & 17.49 & 36.58 & 37.09 & 21.51 & 38.06 & 41.75 & \textbf{43.38} \\
			& 3     & 23.53 & 56.50 & 57.31 & 29.11 & 58.28 & 61.17 & \textbf{61.82} \\
			& 5     & 25.51 & 61.93 & 64.01 & 31.63 & 64.98 & 66.49 & \textbf{67.50} \\
			& 10    & 26.60 & 66.31 & 68.28 & 32.93 & 69.25 & 70.83 & \textbf{72.22} \\
			\midrule
			\multirow{4}[2]{*}{ant} & 1     & 20.42 & 37.73 & 38.38 & 23.78 & 39.35 & 41.32 & \textbf{45.01} \\
			& 3     & 26.17 & 57.85 & 57.90 & 31.45 & 58.87 & 61.29 & \textbf{63.50} \\
			& 5     & 27.47 & 63.40 & 63.87 & 33.35 & 64.84 & 66.80 & \textbf{68.33} \\
			& 10    & 28.83 & 67.58 & 68.58 & 34.91 & 69.55 & 71.13 & \textbf{73.13} \\
			\midrule
			\multirow{4}[2]{*}{batik} & 1     & 16.90 & 34.22 & 34.00 & 20.04 & 34.97 & 35.91 & \textbf{38.82} \\
			& 3     & 21.91 & 51.58 & 52.32 & 26.53 & 53.29 & 55.71 & \textbf{56.27} \\
			& 5     & 23.43 & 57.61 & 59.41 & 28.27 & 60.38 & 61.76 & \textbf{62.66} \\
			& 10    & 24.57 & 62.65 & 64.69 & 29.79 & 65.66 & 66.33 & \textbf{68.62} \\
			\midrule
			\multirow{4}[2]{*}{db4o} & 1     & 18.06 & 34.44 & 34.29 & 21.42 & 35.26 & 35.91 & \textbf{41.41} \\
			& 3     & 23.25 & 54.41 & 54.89 & 28.30 & 55.86 & 57.03 & \textbf{59.44} \\
			& 5     & 24.85 & 60.96 & 61.33 & 30.25 & 62.30 & 63.07 & \textbf{66.44} \\
			& 10    & 26.16 & 65.20 & 66.71 & 31.84 & 67.68 & 68.02 & \textbf{71.72} \\
			\midrule
			\multirow{4}[2]{*}{itext} & 1     & 19.24 & 35.34 & 34.77 & 23.00 & 35.74 & 38.90 & \textbf{41.66} \\
			& 3     & 24.80 & 56.25 & 56.10 & 30.06 & 57.07 & 59.59 & \textbf{62.12} \\
			& 5     & 27.05 & 61.82 & 62.35 & 32.71 & 63.32 & 65.45 & \textbf{67.78} \\
			& 10    & 28.38 & 66.16 & 67.53 & 34.35 & 68.50 & 69.79 & \textbf{72.22} \\
			\midrule
			\multirow{4}[2]{*}{jgit} & 1     & 21.62 & 35.92 & 36.35 & 25.57 & 37.32 & 41.31 & \textbf{45.98} \\
			& 3     & 28.01 & 56.96 & 58.23 & 33.67 & 59.20 & 61.71 & \textbf{65.33} \\
			& 5     & 30.27 & 63.73 & 64.12 & 36.32 & 65.09 & 67.46 & \textbf{70.12} \\
			& 10    & 31.62 & 67.60 & 69.12 & 37.84 & 70.09 & 71.52 & \textbf{74.32} \\
			\midrule
			\multirow{4}[2]{*}{jts} & 1     & 16.20 & 35.60 & 35.01 & 19.78 & 35.98 & 38.18 & \textbf{39.31} \\
			& 3     & 21.11 & 54.50 & 54.41 & 26.08 & 55.38 & 57.42 & \textbf{56.53} \\
			& 5     & 23.01 & 59.40 & 60.29 & 28.34 & 61.26 & 62.50 & \textbf{63.29} \\
			& 10    & 23.94 & 63.47 & 64.62 & 29.55 & 65.59 & 66.88 & \textbf{68.11} \\
			\midrule
			\multirow{4}[2]{*}{maven} & 1     & 18.82 & 37.78 & 39.26 & 22.00 & 40.23 & 43.11 & \textbf{44.35} \\
			& 3     & 23.58 & 57.20 & 58.23 & 28.32 & 59.20 & 61.41 & \textbf{61.71} \\
			& 5     & 25.42 & 61.79 & 63.59 & 30.17 & 64.56 & 65.89 & \textbf{67.15} \\
			& 10    & 26.39 & 65.38 & 67.44 & 31.63 & 68.41 & 69.95 & \textbf{71.68} \\
			\midrule
			\multirow{4}[2]{*}{poi} & 1     & 21.37 & 37.78 & 37.93 & 25.09 & 38.90 & 43.38 & \textbf{46.74} \\
			& 3     & 28.08 & 57.66 & 57.66 & 33.48 & 58.63 & 62.89 & \textbf{65.11} \\
			& 5     & 29.67 & 63.62 & 64.55 & 35.49 & 65.52 & 68.15 & \textbf{69.29} \\
			& 10    & 30.93 & 67.63 & 68.98 & 37.04 & 69.95 & 72.08 & \textbf{73.74} \\
			\bottomrule
		\end{tabular}%
	\end{center}
	\label{Table:Accuracy}
\end{table}%

Further, to see the effectiveness of the proposed approach, we evaluate its performance with the random test set. In random testing the model is trained on one project and then tested with different a project. The random test results are presented in \hyperref[Table:Accuracy-Random]{Table} \ref{Table:Accuracy-Random} where $P_{train}$ shows the project name used for training and $p_{test}$ shows the project name used for testing. On average it improves the accuracy (\textit{K@1}) by \textit{5.66\%} in random test and by \textit{6.62\%} in independent (Cassandra) testing from the best baseline (DNN). From the results, we conclude that the proposed approach outperforms other baseline approaches in both cases, independent (Cassandra) and random testing. 

\begin{table}[htbp]
	\caption{Accuracy comparison of proposed approach with previous works with random test set where $P_{train}$ shows the project name used for training and $p_{test}$ shows the project name used for testing.}
	\begin{center}
		\begin{tabular}{|c|c|c|ccc|cccc|}
			\toprule
			&   &   & \multicolumn{3}{c|}{Previous Works} & \multicolumn{4}{c|}{\textbf{Our Work}} \\
			$P_{train}$ & $P_{test}$  & K  & N-gram & RNN   & DNN   & {N-gram+} & {RNN+} & {GRU} & {CodeGRU} \\
			\bottomrule
			\multirow{4}[2]{*}{antlr} & \multirow{4}[2]{*}{batik}  
			& 1     & 25.20 & 35.72 & 39.29 & 32.41 & 40.25 & 43.27 & \textbf{44.06} \\
			& & 3     & 33.48 & 55.04 & 58.05 & 45.43 & 59.01 & \textbf{63.14} & 62.27 \\
			& & 5     & 36.21 & 60.16 & 64.29 & 48.75 & 65.25 & 68.00 & \textbf{68.06} \\
			& & 10    & 37.80 & 65.03 & 69.15 & 51.18 & 70.12 & \textbf{72.23} & 72.16 \\
			\midrule
			\multirow{4}[2]{*}{ant} & \multirow{4}[2]{*}{db4o}  
			& 1     & 24.50 & 36.63 & 36.85 & 30.48 & 37.82 & 39.45 & \textbf{42.01} \\
			& & 3     & 31.97 & 56.22 & 56.14 & 39.58 & 57.11 & 59.24 & \textbf{59.89} \\
			& & 5     & 34.50 & 61.75 & 62.29 & 42.59 & 63.26 & \textbf{65.52} & 64.40 \\
			& & 10    & 36.40 & 66.62 & 67.78 & 45.35 & 68.74 & 70.30 & \textbf{71.18} \\
			\midrule
			\multirow{4}[2]{*}{batik} & \multirow{4}[2]{*}{ant}  
			& 1     & 33.75 & 40.29 & 43.81 & 42.05 & 44.78 & 44.61 & \textbf{48.12} \\
			& & 3     & 42.83 & 57.10 & 61.47 & 54.92 & 62.43 & 63.39 & \textbf{64.76} \\
			& & 5     & 45.56 & 62.37 & 66.78 & 58.09 & 67.74 & 68.85 & \textbf{70.50} \\
			& & 10    & 47.64 & 67.09 & 71.92 & 61.50 & 72.89 & 73.56 & \textbf{75.89} \\
			\midrule
			\multirow{4}[2]{*}{db4o} & \multirow{4}[2]{*}{jgit}  
			& 1     & 24.21 & 35.52 & 35.93 & 30.56 & 36.90 & 37.94 & \textbf{41.97} \\
			& & 3     & 34.59 & 55.61 & 56.61 & 42.56 & 57.57 & 58.86 & \textbf{60.19} \\
			& & 5     & 36.70 & 61.89 & 62.30 & 59.36 & 63.27 & 64.47 & \textbf{67.01} \\
			& & 10    & 38.74 & 65.88 & 67.39 & 49.36 & 68.35 & 69.28 & \textbf{72.07} \\
			\midrule
			\multirow{4}[2]{*}{itext} & \multirow{4}[2]{*}{jts}  
			& 1     & 24.88 & 36.84 & 39.14 & 33.29 & 40.11 & 41.66 & \textbf{44.14} \\
			& & 3     & 34.03 & 55.81 & 58.61 & 47.58 & 59.57 & 62.29 & \textbf{63.77} \\
			& & 5     & 37.14 & 61.72 & 64.24 & 51.00 & 65.21 & 67.57 & \textbf{68.74} \\
			& & 10    & 38.80 & 66.28 & 68.88 & 53.19 & 69.84 & 71.99 & \textbf{73.14} \\
			\midrule
			\multirow{4}[2]{*}{jgit} & \multirow{4}[2]{*}{itext}  
			& 1     & 27.49 & 35.39 & 38.13 & 33.81 & 39.10 & 41.99 & \textbf{47.79} \\
			& & 3     & 35.12 & 56.00 & 58.26 & 46.75 & 59.22 & 62.52 & \textbf{65.33} \\
			& & 5     & 37.65 & 61.84 & 64.15 & 49.95 & 65.12 & 67.67 & \textbf{70.77} \\
			& & 10    & 39.56 & 66.92 & 69.24 & 52.71 & 70.65 & 72.31 & \textbf{76.32} \\
			\midrule
			\multirow{4}[2]{*}{jts} & \multirow{4}[2]{*}{poi}  
			& 1     & 30.09 & 40.21 & 40.56 & 33.29 & 41.53 & 43.61 & \textbf{44.65} \\
			& & 3     & 40.01 & 57.88 & 58.35 & 48.40 & 59.31 & \textbf{62.07} & 60.89 \\
			& & 5     & 43.50 & 62.72 & 63.47 & 53.08 & 64.44 & 66.67 & \textbf{66.94} \\
			& & 10    & 44.84 & 66.63 & 67.75 & 54.86 & 68.72 & 70.64 & \textbf{71.69} \\
			\midrule
			\multirow{4}[2]{*}{maven} & \multirow{4}[2]{*}{antlr}  
			& 1     & 25.99 & 39.77 & 41.67 & 35.53 & 42.63 & 46.43 & \textbf{47.68} \\
			& & 3     & 32.58 & 58.26 & 60.90 & 45.12 & 61.87 & \textbf{65.40} & 61.25 \\
			& & 5     & 34.95 & 63.66 & 66.13 & 48.22 & 67.87 & \textbf{69.75} & 67.47 \\
			& & 10    & 37.04 & 67.18 & 70.37 & 50.99 & 71.33 & 73.35 & \textbf{73.41} \\
			\midrule
			\multirow{4}[2]{*}{poi} & \multirow{4}[2]{*}{maven}  
			& 1     & 33.12 & 41.42 & 41.26 & 36.11 & 42.22 & 48.17 & \textbf{49.14} \\
			& & 3     & 41.85 & 60.76 & 60.90 & 52.21 & 61.87 & 65.22 & \textbf{66.16} \\
			& & 5     & 44.34 & 65.48 & 66.38 & 55.89 & 67.34 & 69.65 & \textbf{69.66} \\
			& & 10    & 46.37 & 69.15 & 70.30 & 58.25 & 71.27 & 72.90 & \textbf{73.58} \\
			\bottomrule
		\end{tabular}%
	\end{center}
	\label{Table:Accuracy-Random}
\end{table}%

\subsection{Source Code Suggestion and Completion\label{MRRScores}}
To quantify the accuracy of our proposed approach for the source code suggestion and completion task we calculate the \textit{Mean Reciprocal Rank (MRR)}. The MRR is a rank based evaluation metric in which suggestions that occur earlier in the list are weighted higher than those that occur later in the list. The MRR produces a value between \textit{0-1}, where the value \textit{1} indicates perfect source code suggestion model. The MRR can be expressed as

\begin{equation}
MRR(C) = \dfrac{1}{|{C}|}\sum_{i=1}^{|{C}|}\dfrac{1}{y^i}
\end{equation}

where ${C}$ is code sequence and $y^i$ refers to the index of the first relevant prediction. $MRR(C)$ is the average of all sequences $C$ in the test data set.

\hyperref[Table:MRR]{Table} \ref{Table:MRR}  shows the MRR score of independent test set (Cassandra). The MRR score lies between 0-1, and a higher value indicates a better source code suggestion and completion model. CodeGRU achieves the lowest MRR score of \textit{0.447} and the highest MRR score is \textit{0.559}, while previous models lowest MRR score is \textit{0.433} and the highest MRR score is \textit{0.492}. On average proposed approach gains the MRR score of \textit{0.500}, while previous studies only gain \textit{0.465}.

\begin{table}[htbp]
	\caption{Mean Reciprocal Rank comparison of our proposed approach with previous works with independent test set (Cassandra).}
	\begin{center}
		\begin{tabular}{|c|ccc|cccc|}
			\toprule
			& \multicolumn{3}{c|}{Previous Works} & \multicolumn{4}{c|}{\textbf{Our Work}} \\
			& N-Gram & RNN   & DNN   & N-gram+ & RNN+  & GRU   & CodeGRU \\
			\bottomrule
			antlr & 0.209 & 0.467 & 0.479 & 0.258 & 0.486 & 0.517 & \textbf{0.528}
			\\
			\midrule
			ant   & 0.235 & 0.478 & 0.487 & 0.280 & 0.494 & 0.515 & \textbf{0.544} \\
			\midrule
			batik & 0.197 & 0.433 & 0.440 & 0.236 & 0.447 & 0.461 & \textbf{0.481} \\
			\midrule
			db4o  & 0.209 & 0.447 & 0.451 & 0.252 & 0.458 & 0.467 & \textbf{0.511} \\
			\midrule
			itext & 0.225 & 0.457 & 0.458 & 0.270 & 0.465 & 0.494 & \textbf{0.520} \\
			\midrule
			jgit  & 0.252 & 0.467 & 0.477 & 0.301 & 0.484 & 0.516 & \textbf{0.557} \\
			\midrule
			jts   & 0.190 & 0.450 & 0.452 & 0.233 & 0.459 & 0.480 & \textbf{0.485} \\
			\midrule
			maven & 0.215 & 0.478 & 0.492 & 0.255 & 0.499 & 0.523 & \textbf{0.534} \\
			\midrule
			poi   & 0.250 & 0.475 & 0.485 & 0.296 & 0.492 & 0.533 & \textbf{0.559} \\
			\midrule
			\textbf{Average} & 0.220 & 0.461 & 0.469 & 0.265 & 0.476 & 0.501 & \textbf{0.524} \\
			
			\bottomrule
		\end{tabular}%
	\end{center}
	\label{Table:MRR}
\end{table}%

\begin{figure*}[htbp]
	\centering
	\includegraphics[width=\linewidth]{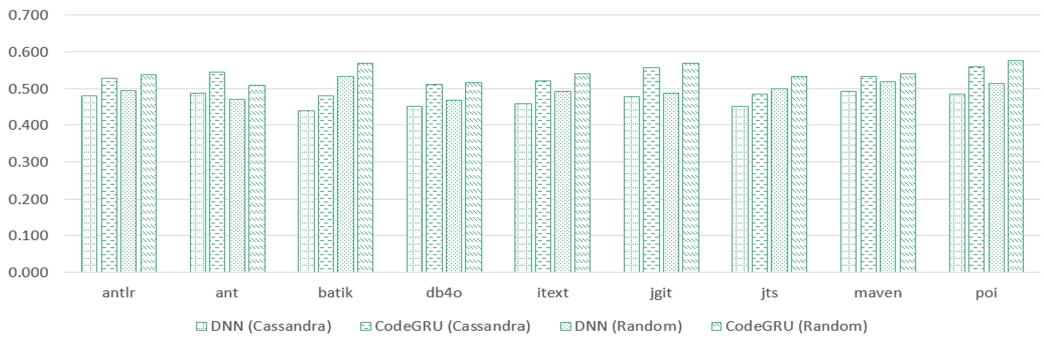}
	\caption{MRR Score Comparison.}
	\label{fig:MRR-Comp}
\end{figure*}

The random test results are presented in \hyperref[Table:MRR-random]{Table} \ref{Table:MRR-random}. From \hyperref[Table:MRR-random]{Table} \ref{Table:MRR-random} and \hyperref[fig:MRR-Comp]{Fig.} \ref{fig:MRR-Comp}, we make the following observations:

\begin{itemize}
	\item The average MRR result of proposed approach is \textit{0.524} and \textit{0.543}, while the best baseline (DNN) average MRR score is \textit{0.469} and \textit{0.497} for independent (Cassandra) and random test respectively. 
	\item From the results, we conclude that the proposed approach outperforms other baseline approaches.
\end{itemize}

\begin{table*}[htbp]
	\caption{Mean Reciprocal Rank comparison of our proposed approach with previous works with random test set where $P_{train}$ shows the project name used for training and $P_{test}$ shows the project name used for testing.}
	\begin{center}
		\begin{tabular}{|c|c|ccc|cccc|}
			\toprule
			& & \multicolumn{3}{c|}{Previous Works} & \multicolumn{4}{c|}{\textbf{Our Work}} \\
			$P_{train}$  & $P_{test}$ & N-Gram & RNN   & DNN   & N-gram+ & RNN+  & GRU   & CodeGRU \\
			\bottomrule
			antlr & batik & 0.298 & 0.455 & 0.495 & 0.395 & 0.501 & 0.533 & \textbf{0.537}\\
			\midrule
			ant   & db4o & 0.286 & 0.465 & 0.471 & 0.356 & 0.478 & 0.496 & \textbf{0.508} \\
			\midrule
			batik & ant & 0.388 & 0.490 & 0.532 & 0.491 & 0.539 & 0.542 & \textbf{0.570} \\
			\midrule
			db4o  & jgit & 0.297 & 0.458 & 0.467 & 0.371 & 0.474 & 0.485 & \textbf{0.515} \\
			\midrule
			itext & jts &  0.299 & 0.493 & 0.493 & 0.410 & 0.499 & 0.521 & \textbf{0.540} \\
			\midrule
			jgit  & itext & 0.319 & 0.458 & 0.487 & 0.410 & 0.494 & 0.526 & \textbf{0.568} \\
			\midrule
			jts   & poi & 0.357 & 0.490 & 0.499 & 0.416 & 0.505 & 0.530 & \textbf{0.532} \\
			\midrule
			maven & antlr & 0.298 & 0.492 & 0.518 & 0.410 & 0.525 & \textbf{0.560} & 0.540 \\
			\midrule
			poi   & maven & 0.380 & 0.510 & 0.515 & 0.444 & 0.521 & 0.568 & \textbf{0.576} \\
			\midrule
			\multicolumn{2}{|c|}{\textbf{Average}}  & 0.326 & 0.476 & 0.497 & 0.411 & 0.504 & 0.529 & \textbf{0.543} \\
			\bottomrule
		\end{tabular}%
	\end{center}
	\label{Table:MRR-random}
\end{table*}%

We conduct ANOVA to measure the statistical significant difference between the proposed approach and the best baseline (DNN). ANOVA is conducted on MRR, where the unit of analysis is a project. We conduct ANOVA on Microsoft Excel version 2016 with its default setting, and no adjustments are involved. \hyperref[Table:ANOVA]{Table} \ref{Table:ANOVA} shows\textit{ F $>$ F-crit} and \textit{p-value $<$ (alpha = 0.05)} are true for MRR in both cases (Cassandra and Random test); therefore, we reject the null hypothesis, suggesting that a statistically significant difference exist.

\begin{table}[htbp]
	\small
	\centering
	\caption{ANOVA Analysis on MRR Scores}
	\begin{tabular}{lrrrrrr}
		\toprule
		\multicolumn{1}{c}{\textit{Source }} & \multicolumn{1}{c}{\textit{SS}} & \multicolumn{1}{c}{\textit{df}} & \multicolumn{1}{c}{\textit{MS}} & \multicolumn{1}{c}{\textit{F}} & \multicolumn{1}{c}{\textit{P-value}} & \multicolumn{1}{c}{\textit{F crit}} \\
		\midrule
		\multicolumn{1}{c}{\textit{MRR (Cassandra)}} &       &       &       &       &       &  \\
		Between Groups & 0.013661 & 1     & 0.013661 & 23.71224562 & 0.00017043 & 4.493998 \\
		Within Groups & 0.009218 & 16    & 0.000576 &       &       &  \\
		Total & 0.022879 & 17    &       &       &       &  \\
		&       &       &       &       &       &  \\
		\multicolumn{1}{c}{\textit{MRR (Random)}} &       &       &       &       &       &  \\
		Between Groups & 0.009253 & 1     & 0.009253 & 17.99466376 & 0.00062147 & 4.493998 \\
		Within Groups & 0.008227 & 16    & 0.000514 &       &       &  \\
		Total & 0.01748 & 17    &       &       &       &  \\
		&       &       &       &       &       &  \\
		\bottomrule
	\end{tabular}%
	\label{Table:ANOVA}%
	\\
	Where, SS = sum of squares, df = degree of freedom, MS = mean square.
\end{table}%

\subsection{Impact of \textit{CodeGRU} on Vocabulary}
NLP based deep learning models suffer from the vocabulary size. Usually, they are trained on a fixed size vocabulary and replace the out of vocabulary tokens with some special token while testing. The same approach has been employed by previous studies \cite{hindle2012naturalness,white2015toward,hindle2012naturalness}. In this work, we employ a novel approach of capturing the token types which helps reduce the vocabulary size up to 24.93\% which suggest that our approach can help minimize the out of vocabulary issue. \hyperref[Table:Vocabulary]{Table} \ref{Table:Vocabulary} shows the vocabulary statistic with and without our proposed approach. By capturing the token type information the vocabulary size reduces significantly, in some cases over \textit{30\%}. Vocabulary size reduction helps overcome two limitations; It helps minimize the out of vocabulary issue and reduce the time, and computation cost for model training. It took approximately 4-6 hours to train the \textit{jts} model by using the proposed approach whereas without our proposed approach it took 8-20 hours for the same project.

\begin{table}[htbp]
	\caption{Vocabulary comparison with and without our proposed approach.}
	\begin{center}
		\begin{tabular}{|c|ccc|c|}
			\toprule
			{Projects} & {Code Tokens} & {$V_{Norm}$} & {$V_{CodeGRU}$} & {\% Decrease}\\
			\bottomrule
			ant        & 920,978 	&  17,132 	& 12,417 	& \textbf{27.52\%}\\
			\midrule
			cassandra  & 2,734,218 &  33,424 	& 26,960 	& \textbf{19.34\%}\\
			\midrule
			db40       & 1,435,382 &  20,286 	& 16,397 	& \textbf{19.17\%}\\
			\midrule
			jgit       & 1,538,905 &  20,970 	& 16,433 	& \textbf{21.64\%}\\
			\midrule
			poi        & 2,876,253 &  47,756 	& 33,867 	& \textbf{29.08\%}\\
			\midrule
			maven      & 494,379 	&  8,066 	& 6,770 	& \textbf{16.07\%}\\
			\midrule
			batik      & 1,246,157 &  21,964 	& 14,643 	& \textbf{33.33\%}\\
			\midrule
			jts        & 611,392 	&  11,903 	& 7,710 	& \textbf{35.23\%}\\
			\midrule
			itext      & 1,164,362 &  19,113 	& 12,824 	& \textbf{32.90\%}\\
			\midrule
			antlr      & 407,248 	&  6,813 	& 5,566 	& \textbf{18.30\%}\\
			\bottomrule
			{Total}	& {13,429,274} & {177,342} 	& {133,135}	& \textbf{24.93\%}\\
			\bottomrule
		\end{tabular}%
	\end{center}
	\label{Table:Vocabulary}
\end{table}%

\subsection{Improving the Performance of CodeGRU}
\begin{figure*}[htbp]
	\centering
	\includegraphics[width=0.8\linewidth]{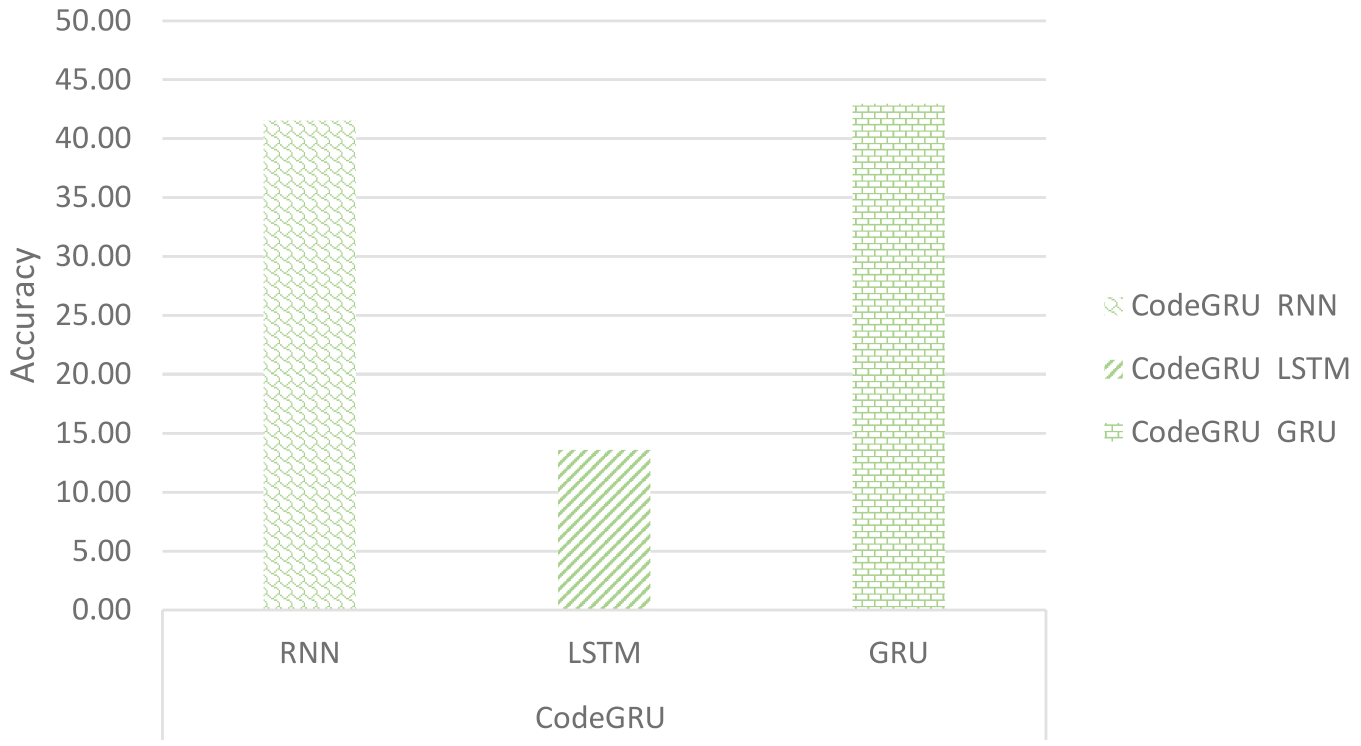}
	\caption{Performance of CodeGRU with different recurrent neural architectures.}
	\label{fig:CodeGRU-Arch-Comp}
\end{figure*}

We conduct an experiment to study the impact of different hyper parameters on CodeGRU performance. We tested various model settings varying the hidden units (\textit{100,200,300}) with the context size of (\textit{10,15,20}) and optimizer (\textit{Adam, RMSprop}). In our experiments, we found that CodeGRU performs well when the hidden units are \textit{300} and context size is \textit{20}. The high values of both parameters (context size and hidden units) are not surprising and most commonly used in various source code modeling tasks \cite{white2015toward,santos2018syntax}. The value of context size (20) reflects that the model is capable of effectively learning the long-term context dependencies of the source code by utilizing 300 hidden units(neurons) whereas smaller context size or hidden units cause the model to under-fit. Further, to alleviate the issue of overfitting, we have adopted dropout regularization which helps prevent the model from overfitting. We choose the code embedding size of \textit{300} to match the hidden units size. Further, in our experiments we found that \textit{Adam} optimizer performed well as compared to \textit{RMSprop} optimizer with the learn rate set to its default \textit{0.001} in both cases. We conclude from the experiments that the architecture matching Table \ref{Table:DeepModelsArch} performed well as compared to other settings. Furthermore, we empirically selected GRU over RNN and LSTM because of its good performance. GRU is an advanced version of LSTM which works similar to it but performs better \cite{tay2018recurrently} by exposing whole hidden context. We train RNN, LSTM and GRU based models by adopting the proposed approach. \hyperref[fig:CodeGRU-Arch-Comp]{Fig.} \ref{fig:CodeGRU-Arch-Comp} shows the average accuracy (K@1) score of each model with independent test set. The performance of RNN and GRU is fairly similar. However, RNN suffers from the vanishing gradient issue which can be overcome by using GRU. Further, GRU exposes the whole hidden context on each time step which is ideal for source code modeling tasks where context dependencies are separated far apart. From the results, we can perceive that the GRU based approach performs well as compared to others.

\subsection{Time and memory Cost}
While conducting our experiments we retain time and space cost of using \textit{Code Analyzer}. \hyperref[Table:time]{Table} \ref{Table:time} shows the time and space cost of each project. For other processes, including tokenization, building sequences, models training, and testing are all common procedures, so we do not analyze their costs. Among all the projects, the time and space cost varies from 4.59 seconds with memory cost of 2.29MB for \textit{antlr} to 32.0 seconds with memory cost of 16.3MB for \textit{poi}.
\begin{table}[htbp]
	\caption{Time and space cost of Code Analyzer.}
	\begin{center}
		\begin{tabular}{|c|c|c|}
			\toprule
			\textbf{Project}	& \textbf{Time(sec)} & \textbf{Space(MB)} \\
			\bottomrule
			antlr & 4.59  & 2.29 \\
			ant   & 10.55 & 4.93 \\
			batik & 14.66 & 7.08 \\
			cassandra & 30.4  & 15.7 \\
			db4o  & 17.37 & 7.58 \\
			itext & 13.81 & 6.25 \\
			jgit  & 16.7  & 8.37 \\
			jts   & 6.94  & 3.44 \\
			maven & 5.98  & 3.04 \\
			poi   & 32.3  & 16.3 \\
			\bottomrule
		\end{tabular}%
	\end{center}
	\label{Table:time}
\end{table}%

\section{Applications of CodeGRU \label{Usecase}} 
In this section, we discuss CodeGRU with two software engineering applications: code suggestion, which aims to suggest multiple predictions for the next code token, and code completion, which aims to complete the whole next source code sequence. We use Visual Studio Code (VSC) IDE version \textit{1.31.1} for the demonstration purpose. The demonstration is conducted on Intel(R) Core(TM) i5-6500 \@ 3.20GHz with 4 cores and 8GB of ram running windows 10 operating system. The CodeGRU takes less than 260MB memory space for the prediction purpose with the largest subject project (POI). We use CodeGRU to help predict the top-k suggestions given a source code context and then use the type information from the IDE to help suggest the identifier names as discussed earlier in section \ref{CodeGRUInDetail}.

\subsection{Code Suggestion \label{CodeSuggestion}} 
\begin{figure}[!ht]
	\subfloat[\label{subfig-1:CodeSuggestionExample}]{%
		\includegraphics[width=\textwidth]{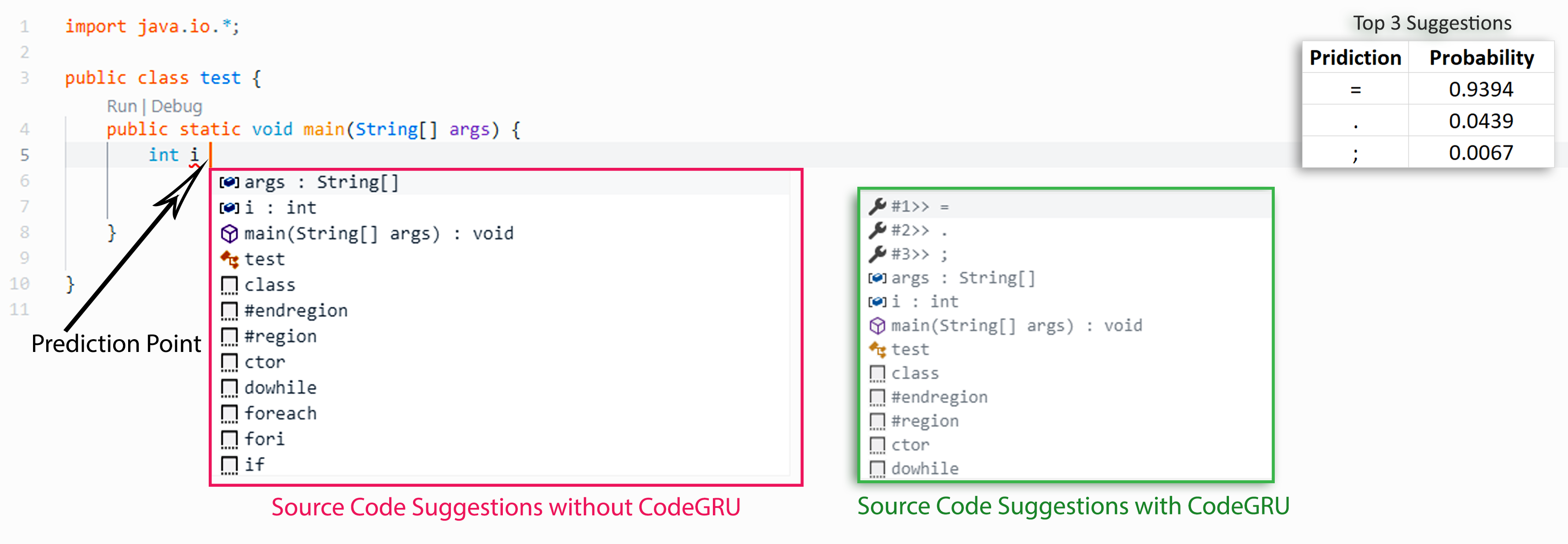}
	}
	\hfill
	\subfloat[\label{subfig-2:CodeSuggestionExample}]{%
		\includegraphics[width=\textwidth]{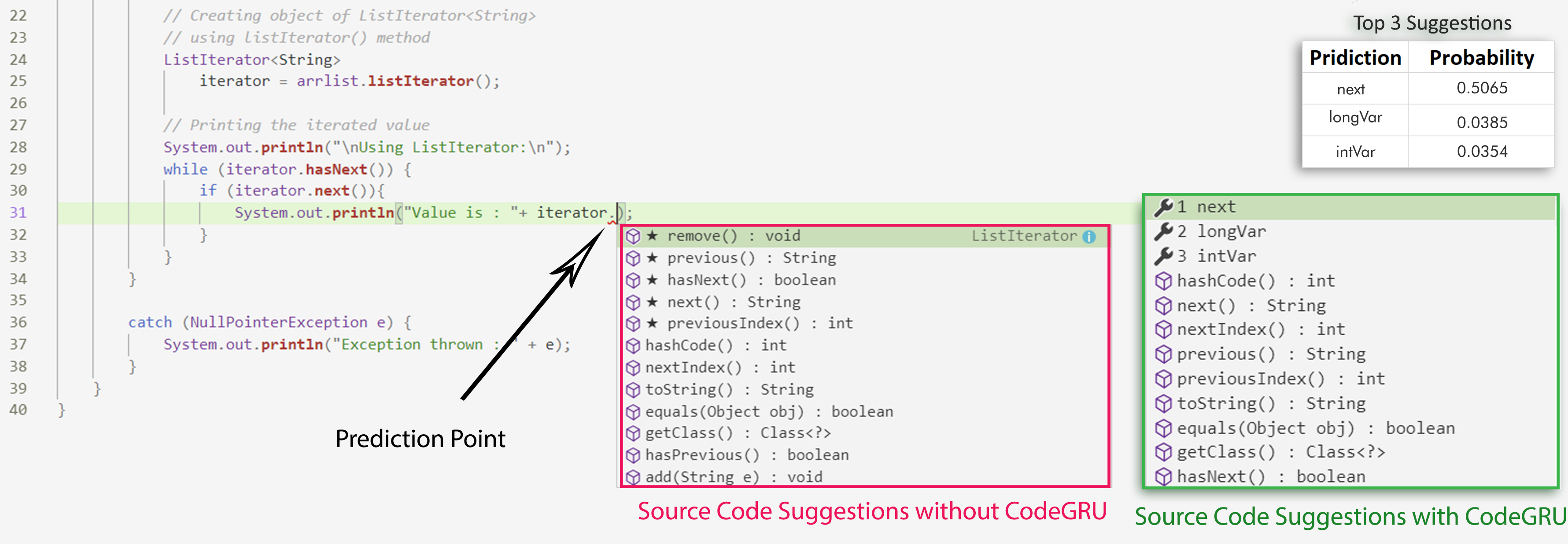}
	}
	\caption{Source Code suggestion example in visual studio code IDE.}
	\label{fig:CodeSuggestionExample}
\end{figure}

CodeGRU is capable of ranking the next code token suggestions by calculating the likelihood based on a given source code context. In \hyperref[subfig-1:CodeSuggestionExample]{Fig.} \ref{subfig-1:CodeSuggestionExample} one can see an example for source code suggestion task where a software developer is defining a variable at line five and the most probable next source code token can be \inlinecode{=} but in visual studio code it does not provide any relevant suggestion whereas CodeGRU suggests the correct next code token at first position of its suggestion list. Consider a more complex example \hyperref[subfig-2:CodeSuggestionExample]{Fig.} \ref{subfig-2:CodeSuggestionExample} where a software developer is about to print a \inlinecode{ListIterator} item at line 31. We can see from the given context the CodeGRU suggest the possible next source code token at its first index whereas visual studio code rank it on its 4th index. In \hyperref[fig:CodeSuggestionExample]{Fig.} \ref{fig:CodeSuggestionExample} one can see CodeGRU suggest the next code token in its top three suggestion list.

\subsection{Code Completion} \label{CodeCompletion}
\begin{figure*}[h]
	\centering
	\includegraphics[width=\linewidth]{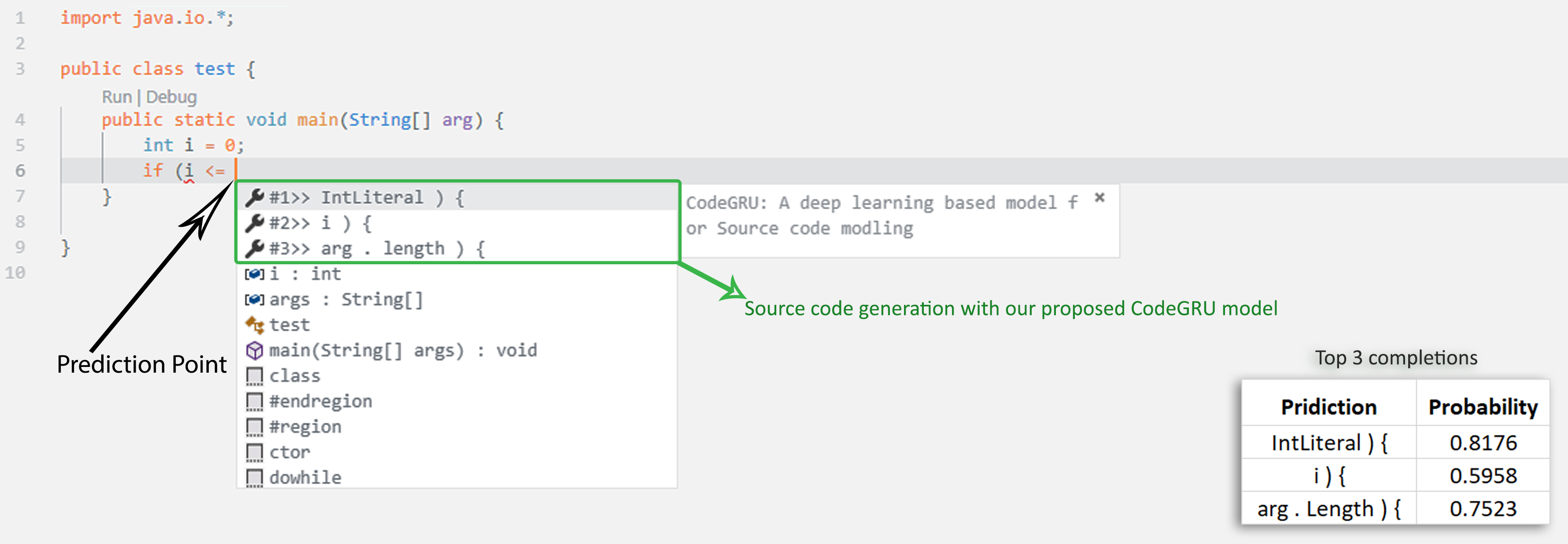}
	\caption{Source code completion example.}
	\label{fig:CodeCompletionExample}
\end{figure*}

The CodeGRU is capable of completing the whole sequence of code with correct syntax and context. Lets take the example discussed in \hyperref[fig:CodeCompletionExample]{Fig.} \ref{fig:CodeCompletionExample} where a developer is writing source code \inlinecode{if ( i <= } at line \textit{6} in visual studio code. As \hyperref[fig:CodeCompletionExample]{Fig.} \ref{fig:CodeCompletionExample} shows CodeGRU provides the top-3 possible code completions. We can observe that CodeGRU capture the data types of the identifiers \inlinecode{i} and \inlinecode{arg} and assign the highest probabilities to the sequence \inlinecode{if ( i <= IntLiteral ) \{} and \inlinecode{if ( i <= arg . Length ) \{}. We can also observe that it give low probability to \inlinecode{if ( i <= i )\{} which is the most unlikely sequence given the context. 

\subsection{Utility Applications}
The application of the proposed approach is not limited to the preceding ones. Usually in the modeling of source code identifiers are removed or replaced by some generic code token (e.g. \textit{idf, unk, etc.}) which may present vital information to help improve the model's performance. The proposed approach can help improve the performance of such approaches \cite{mi2018improving,santos2018syntax,wang2016automatically,mou2016convolutional,xiao2019improving} by considering the token type information. One possible application of our work can be syntax error correction similar to Santos et al \cite{santos2018syntax}. They present NLP based recurrent neural model to detect and correct syntax errors. Their work considers the single token syntax error (113 syntax token) and replacing the identifies with a generic token. Another possible application of our approach can be bug prediction similar to Wang et al. \cite{wang2016automatically}. Our proposed approach may help improve these works \cite{mi2018improving,santos2018syntax,wang2016automatically,mou2016convolutional,xiao2019improving} by capturing the token type information.

\section{Limitations and Threats to Validity}
\subsection{Limitations:} One of the limitations of NLP based models is vocabulary size. Vocabulary size reduction helps in many ways, first, it helps reduce the train time of the model. Second, it helps reduce the computational complexity. Larger the vocabulary size of a model, more time and computational power it will require to train the model. Although our approach reduces the vocabulary size by 24.93\% by capturing the token type information, still there may be some identifier types which may not be captured while training the model. This limitation can be overcome by training the model with even larger data set, but it will increase the vocabulary size which will increase the computational complexity. A solution to this problem is to use char level encoding. Another limitation of this work is the gap concerning the recommendation on the identifier type and the possible identifier names that developers may pick for the source code suggestion task. In this work, we capture the source code context by capturing the token type information and fill the identifier names based on the captured token types. This approach may possibly overload the developer when there are a large number of possible initializations for a token type. In our future work, we consider to overcome these limitations and to provide a more robust approach.  

\subsection{Construct Validity:}  All models are developed using \textit{keras} version \textit{2.2.4} with \textit{tensorflow} version \textit{1.13.1} backend. Although our experiments are detailed and results have shown the effectiveness of our approach but still neural networks are in its infancy. Change in neural network settings, training or evaluating with a different test set may produce different results. Another threat to construct validity is the suitability of the evaluation metrics. The Top-k \cite{white2015toward,hindle2012naturalness,nguyen2018deep} metric is commonly used for the evaluation of deep learning based source code models. However, the proposed approach is further evaluated with MRR \cite{nguyen2018deep,santos2018syntax} and ANOVA metrics to show its effectiveness and to alleviate this threat.

\subsection{Internal Validity:} Our tool suggests the most relevant suggestions provided by the language model given a source code context but the choice of identifier names is still up to the developer. Our tool will suggest the token type for an identifier and actual code token for the rest of the code tokens. Further, We have adopted a beam search approach for the source code completion task. The completion provided by CodeGRU model may not be accurate in some cases as it selects the first prediction for each next token where the correct source code token may not be on the first index. Another threat to internal validity is the implementation of the baselines. We follow the same process as described in the original manuscript of baselines. To alleviate this threat we double checked the implementation and the  results; However, there could be unnoticed imperfections.

\subsection{External Validity:} A threat to external validity is that the generalization of our results. The data set used in this study was collected from \textit{GitHub}, a well-known source code repositories provider. It is not necessary that the projects used in this study represent Java language or other languages code entirely. Another threat to external validity is the choice of hyper-parameters. There is no universal approach to learn the best model parameters, thus the parameter tuning is mainly empirical.

\section{Related Work} \label{RelatedWork}
Most of the modern IDEs provide code completion and code suggestion features. In recent years, deep neural techniques have been successfully applied to various tasks in natural language processing, and also have shown its effectiveness to problems such as code completion, code suggestion, code generation, API mining, code migration, and code categorization. In this section, we discuss prior works which are relevant to this research.

Hindle et al. \cite{hindle2012naturalness} have shown how natural language processing techniques can help in source code modeling. They provide a \textit{n-gram} based model which helps predict the next code token in \textit{Eclipse IDE}. Raychev et al. \cite{raychev2014code} used statistical language model for synthesizing code completions. They applied \textit{n-gram} and \textit{RNN} language model for the task of synthesizing code. Tu et al. \cite{tu2014localness}, proposed a cache based language model that consists of an \textit{n-gram} and a \textit{cache}. Hellendoorn et al. \cite{hellendoorn2017deep} further improved the cache based model by introducing nested locality. White et al. \cite{white2015toward} applied deep learning for source code modeling purpose. Another approach for source code modeling is to use probabilistic context-free grammars(PCFGs) \cite{bielik2016phog}. Allamanis et al. \cite{allamanis2014mining} used a PCFG based model to mine idioms from source code. Maddison et al. \cite{maddison2014structured} used a structured generative model for source code. They evaluated their approach with \textit{n-gram} and \textit{PCFG} based language models and showed how they can help in source code generation tasks. Raychev et al.\cite{raychev2016learning} applied decision trees for predicting API elements. Chan et al. \cite{chen2016similartech} used a graph-based search approach to search and recommend API usages.

Recently there has been an increase in API usage \cite{wang2013mining,keivanloo2014spotting,d2016collective} mining and suggestion. Thung et al. \cite{thung2013automatic} introduced a recommendation system for API methods recommendation by using feature requests. Nguyen et al. \cite{nguyen2016learning} proposed a methodology to learn API usages from byte code. Allamanis et al. \cite{allamanis2014mining} introduced a model which automatically mines source code idioms. A neural probabilistic language model introduced in \cite{allamanis2015suggesting} that can suggest names for the methods and classes. Franks et al. \cite{franks2015cacheca} created a tool for Eclipse named \textit{CACHECA} for source code suggestion using a \textit{n-gram} model. Nguyen et al. \cite{nguyen2012graph} introduced an \textit{Eclipse plugin} which provide context-sensitive code completion based on API usage patterns mining techniques. Chen et al. \cite{chen2016similartech} created a web-based tool to find analogical libraries for different languages. Wang et al. \cite{wang2016automatically} proposed a Deep Belief Network (DBN) based model for inter and cross-project bug prediction. They use source code's AST to learn the semantic features of source code.  Mou et al. \cite{mou2016convolutional} proposed a Convolutional Neural Networks over tree structures to capture source code structural information. They show the effectiveness of their approach for the task of categorizing source code programs based on their functionality. Both works \cite{wang2016automatically,mou2016convolutional} considers certain parts of AST and remove the rest which present vital information for other tasks such as Source code suggestion, code completion etc.

Yin et al. \cite{yin2017syntactic} have proposed a source code generation approach that serially apply actions from a grammar model to generate an abstract syntax tree. A similar work conducted by Rabinovich et al. \cite{rabinovich2017abstract}, which introduced an abstract syntax networks modeling framework for tasks like code generation and semantic parsing. Sethi et al. \cite{sethi2018dlpaper2code} introduced a model which automatically generate source code from deep Learning based research papers. Allamanis et al. \cite{allamanis2015bimodal} proposed a bimodal to help suggest source code snippets with a natural language query. It is also capable of retrieving natural language descriptions with a source code query. Recently deep learning based approaches have widely been applied for source code modeling. Such as code summarization \cite{iyer2016summarizing,allamanis2016convolutional,guerrouj2015leveraging}, code mining \cite{xie2006mapo}, clone detection \cite{kumar2015code},  API learning \cite{gu2016deep} etc. Different from the Santos et al. \cite{santos2018syntax} and Gupta et al. \cite{gupta2017deepfix}, our work focuses on source code suggestion tasks, whereas their works focus on fixing syntax errors. Their work requires the compilation of code after making a fix whereas our work does not require compilation of source code and can provide suggestions instantly.

Our work is relevant to the \cite{hindle2012naturalness,white2015toward,raychev2014code,nguyen2018deep} works, however, it varies from them in several important ways. Hindle et al. \cite{hindle2012naturalness} have shown that source codes are natural and a simple SLM (n-gram) based model can capture the regularities in them. White et al. \cite{white2015toward} and Raychev et al. \cite{raychev2014code} have shown that the neural network based models can capture the regularities much more effectively than n-gram \cite{white2015toward} based models. They applied \textit{RNN} based model to show how deep learning can help improve source code modeling. However, these works \cite{hindle2012naturalness,white2015toward,raychev2014code} consider source code as simple tokens of text whereas our work considers the source code contextual, syntactical and structural information. Further, they treat source code as a single sequence of text tokens with fixed size context window, while we employed a novel variable size context learning approach which shows improvement in the modeling of source code. The most similar work to ours is DNN \cite{nguyen2018deep} however, it varies in several important ways. First, They apply deep neural networks for source code modeling with a fixed size of context. Their work considers the context size of \textit{n=4}, where larger size may cause scalability problem as mentioned in their work \cite{nguyen2018deep}, while our work employed a novel approach of variable size context learning with the upper bound limit of 20 tokens. Further, their approach is limited to Java language, while our work adopts a general approach which can work with different static type languages. Finally, they train and test the model on the same project by splitting each project into ten folds from which one fold is used for test purpose and rest are used for training purpose. Our model is trained on one project and tested on a separate independent project which shows the proposed approach is capable of predicting cross-project source code tokens.

\section{Conclusion \label{Conclusion}} 
This paper presented CodeGRU, a novel approach for source code modeling which captures source code's contextual, structural and syntactical information. The work proposed a novel approach which can capture the source code context by leveraging the token type information. The CodeGRU can effectively capture the right context even it is separated far apart in the code. CodeGRU further proposed a novel approach which learns variable size context while modeling the source code. The evaluation has shown that CodeGRU outperforms the state-of-the-art language models and help reduce the vocabulary size up to 24.93\% which suggest it can help minimize the out of vocabulary issue. We further evaluated CodeGRU with two software engineering applications: (1) source code suggestion, which can suggest multiple predictions for the next code token, and (2) code completion, which can complete whole next code sequence which shows that it is of practical use.

In the future, we would like to evaluate our approach for the dynamic typed languages. In dynamic type languages, a source code token can have different types which makes it difficult to capture the right context. We also aim at providing an end to end solution with a large data set which can help software developers directly utilize these models for both static and dynamic typed languages. Another limitation of deep learning based approaches is computation power, where training a new model require additional resources. A common software developer cannot afford to have a server or GPU based system to train these models. There is a need for centralizing these language models which can directly benefit software developers with minimum effort.

\section*{Acknowledgments}
This work was supported by the National Key R\&D (grant no. 2018YFB1003902), Natural Science Foundation of Jiangsu Province (No. BK20170809), National Natural Science Foundation of China (No. 61972197) and Qing Lan Project.

\bibliography{CodeGRU}

\end{document}